\documentclass[final,times,12pt]{elsarticle}
\usepackage[top=2.4cm,bottom=2.4cm,left=2cm,right=2cm]{geometry}

\usepackage{amsmath}
\usepackage{graphicx}
\usepackage{balance}
\usepackage{comment}
\usepackage{setspace}

\usepackage[table,xcdraw,svgnames]{xcolor}
\usepackage[colorlinks]{hyperref}
\AtBeginDocument{%
 \hypersetup{
   citecolor=SteelBlue,
   linkcolor=SteelBlue,   
   urlcolor=SteelBlue}
  }

\usepackage{lineno}

\biboptions{sort&compress}

\makeatletter
\def\ps@pprintTitle{%
 \let\@oddhead\@empty
 \let\@evenhead\@empty
 \def\@oddfoot{}%
 \let\@evenfoot\@oddfoot}
 \makeatother
 
\usepackage{enumitem}
\setlist[description]{labelindent=25pt,style=multiline,leftmargin=4.5cm}

\begin{document}
\begin{frontmatter}

\title{Uncovering Feature Interdependencies in High-Noise Environments with Stepwise Lookahead Decision Forests}

\author[add1]{Delilah Donick \fnref{EC} } 
\author[add2,add3]{Sandro Claudio Lera \fnref{EC} \corref{CSA}}

\address[add1]{\scriptsize Vencera Consulting, New York, USA}
\address[add2]{\scriptsize Massachusetts Institute of Technology, Cambridge, USA}
\address[add3]{\scriptsize Southern University of Science and Technology, Shenzhen, China}

\fntext[EC]{both authors contributed to this work equally}
\cortext[CSA]{corresponding author (slera@mit.edu)}

\begin{abstract}

Conventionally, random forests are built from ``greedy" decision trees which each consider only one split at a time during their construction.
{
The sub-optimality of greedy implementation has been well-known, yet mainstream adoption of more sophisticated tree building algorithms has been lacking.
We examine under what circumstances an implementation of less greedy decision trees actually yields outperformance. 
To this end, a ``stepwise lookahead" variation of the random forest algorithm is presented for its ability to better uncover binary feature interdependencies.  
}
In contrast {to the greedy approach}, the decision trees included in this random forest algorithm, each simultaneously consider three split nodes in tiers of depth two.
{
It is demonstrated on synthetic data and financial price time series that the lookahead version significantly outperforms the greedy one when
(a) certain non-linear relationships between feature-pairs are present
and (b) if the signal-to-noise ratio is particularly low. 
} 
A long-short trading strategy for copper futures is then backtested by training both greedy and {stepwise lookahead} random forests to predict the signs of daily price returns. 
The resulting superior performance of the lookahead algorithm is at least partially explained by the presence of ``XOR-like'' relationships between long-term and short-term technical indicators.  
More generally, across all examined datasets, when no such relationships between features are present, performance across random forests is similar. 
Given its enhanced ability to understand the feature-interdependencies present in complex systems, this lookahead variation is a useful extension to the toolkit of data scientists, 
{in particular for financial machine learning, where conditions (a) and (b) are typically met.} 

\end{abstract}

\end{frontmatter}

\begin{center}
\begin{figure}[h!]
   \includegraphics[width=\linewidth]{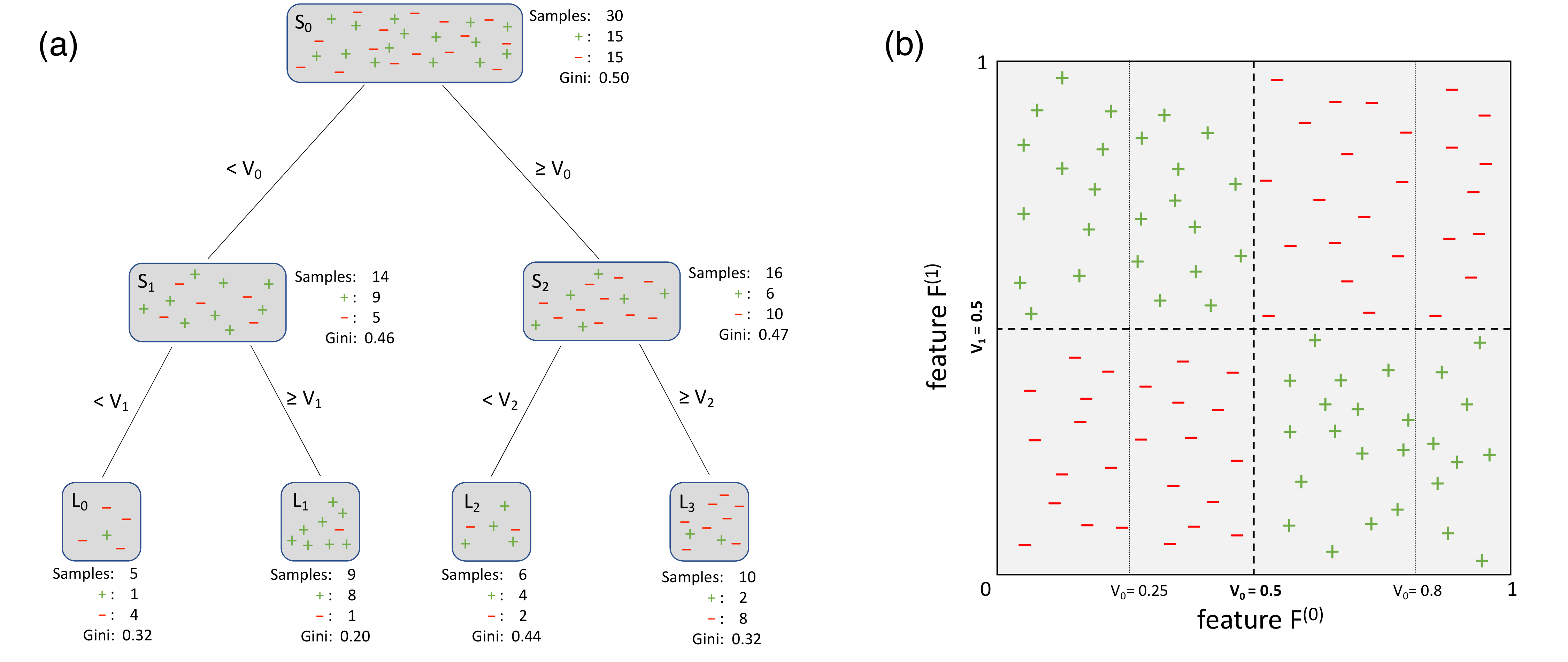}
   \caption{ 
(a)
This DT of depth 2 has split nodes $S_0$, $S_1$, $S_2$,  split features $F^{(0)}$, $F^{(1)}$, $F^{(2)}$, split feature values $V_0$, $V_1$, and $V_2$, and resulting leaf nodes $L_0$, $L_1$, $L_2$, $L_3$. 
The green ``$+$" (red ``$-$'') within the nodes represent training samples that belong to the ``$+$" (``$-$") class. 
All $N=30$ datapoints are contained in the top (root) node.  
If a datapoint has a value for feature $F^{(0)}$ that is greater than or equal to $V_0$, it travels down the right branch into $S_2$.
If not, it travels down the left branch into $S_1$.
This process is then repeated so that samples are further separated into  $L_0$, $L_1$, $L_2$, and $L_3$.
With each split, the samples are separated so that the resulting nodes are each more homogenous than the previous one.
This can be observed qualitatively by looking at the ratio of ``$+$"s to ``$-$"s, or quantitatively by the Gini-scores \eqref{eq:GiniScore} for the nodes at each level of the DT. 
(b)
We show 100 datapoints that follow an XOR-like pattern.
Each datapoint is plotted by its values for Feature $F^{(0)}$ and Feature $F^{(1)}$.
The green ``$+$'"s and the red ``$-$"s denote the associated class labels.
All datapoints which have $F^{(0)}$ and $F^{(1)}$ values that are both either greater or less than 0.5 belong to the ``$-$" class. 
All others are part of the ``$+$'' class. 
The darker, dashed lines at $V_0=0.5$ and $V_1=0.5$ together represent the optimal splits and a perfect in-sample classification of the data.
The lighter, dotted lines at  $V_0=0.25$ and  $V_0=0.8$ represent suboptimal splits which, if selected, would result in a lower cumulative Gini-score \eqref{eq:NGTotalGini}. 
}
\label{fig:LRF_example}
\end{figure}
\end{center}

Along with the rising popularity of machine learning, the random forest (RF) \cite{Breiman2001} has become a ubiquitous algorithm in data science.
Indeed, an examination across a large variety of data sets has shown that the RF outperforms, on average, most other common machine learning methods \cite{Fernandez2014}. 
Its success can be attributed to its ability to handle a variety of tasks with few parameters to tune, 
its ability to accommodate small sample sizes and high-dimensional feature spaces, 
its robustness to noise, 
and its aptness for parallelization\cite{Biau2016}. 
These advantages are a result of the RF's design, as it is an ensemble of decision trees (DTs), popular structures for categorizing and sorting data
(see Figure \ref{fig:LRF_example} (a) for an example).
Another advantage of tree based methods is their interpretability, especially through the recent evaluation techniques developed around Shapley values \cite{Lundberg2020}, 
which readily provide insights into the directional contribution of each feature. 

Given the prevalence of tree-based methods in data science tasks, it is then crucial to be aware of the different types of DTs that may be used. 
The DT was introduced in the 1980s \cite{Breiman1984}, and has since been extrapolated in numerous ways. 
A DT is conventionally constructed in a ``greedy", ``myopic'', top-down fashion \cite{Quinlan1986,Quinlan1993}. 
Starting from the root node, a split is determined by maximizing a homogeneity measure, irrespective of subsequent splits on the two resulting child nodes.
This myopic nature of most of the decision tree thus ranks possible attributes based on their immediate descendants. 
Such a strategy prefers tests that score high in isolation and may overlook combinations of attributes. 
A typical example is two features that form an XOR-junction, shown in Figure \ref{fig:LRF_example} (b) and further elaborated on in the methods section below. 
In such a case, knowing the values of both attributes determines the value of the target. 
However, knowing the value of either attributes does not help at all in predicating the value of the traget.
Thus, a myopic algorithm might assume that neither attributes are useful, and therefore disregard them \cite{Rokach2016}. 
In the remainder of this article, we shall use the terms ``myopic'' and ``greedy'' interchangeably. 
We refer to these greedy decision trees as GDTs. 
Random forests built out of GDTs are referred to as GRFs. 

The concern that GDTs are suboptimal was addressed long ago \cite{Bennett1994}. 
The problem of constructing a globally optimal DT is NP-hard \cite{Laurent1976}, 
Hence, various optimization techniques,
relying on
linear programming \cite{Bennett1994,Bertsimas2017,Verwer2019}, 
stochastic gradient descent \cite{Norouzi2015}, 
mixed-integer formulation \cite{Uney2006},
{anytime induction \cite{Esmeir2004}}, 
randomization \cite{Blanquero2020}, 
multilayer cascade structures \cite{Zhou2017}, 
column generation techniques \cite{Firat2018}, 
and genetic algorithms \cite{Barros2015},
have been proposed to solve this problem. 
All of these methods seek to strike a balance between accuracy, simplicity and efficiency. 
We refer to such optimal, or at least less-greedy, decision trees as lookahead decision trees (LDTs), since they take into consideration a few or all potential future split-points. 
Accordingly, we label random forests constructed from LDTs as LRFs. 

Despite the various approaches already taken towards LDT construction, adaption among main-stream practitioners has been lacking. 
State-of-the-art machine-learning libraries \cite{Pedregosa2011,Chen2016,Ke2017} still rely on greedy heuristics such as the CART \cite{Breiman1984} or ID3 algorithm \cite{Quinlan1986}.
This is, however, only partially explained by computational aspects. 
Another major reason for the lacking adoption of LDTs is that they have failed to consistently outperform their greedy counterparts \cite{Murthy1995,Esmeir2004,Rokach2016}. 
But if the LDTs ought to be better in theory, how can we explain this lack of outperformance? 

In this article, we shine light on this issue by demonstrating that the outperformance of less-greedy decision trees is more prevalent in environments that are characterized by a particularly low signal-to-noise ratio (SNR). 
We find that only in the presence of specific non-linear patterns (such as the XOR-junction) and with sufficiently high noise do the non-greedy decision trees demonstrate superior performance. 
By contrast, many of the standard machine learning test sets are comparatively high signal-to-noise examples.
As our examples on synthetic data show, outperformance of LDTs over GDTs is confined to regimes close to the limit of random guessing. 

As a practical application, we will focus on financial markets for two primary reasons. 
First, financial markets are known to exhibit a particularly unfavorable signal-to-noise ratio and future price returns are thus notoriously hard to predict \cite{dePrado2018}. 
Second, financial markets are hall-mark examples of complex systems, which are characterized by the fact that ``the whole is more than the sum of its parts'' \cite{Anderson1972}.  
Accordingly, the non-greedy examination of multiple parts, or features, in concert, rather than the greedy examination of each part, or feature, in silo, may produce more accurate predictions. 
We specifically focus on an example using copper futures price data, in which we demonstrate that a strategy based on less myopic DTs significantly outperforms a strategy based on greedy ones. 

The remainder of this article is structured as follows: 
In the Methods section, we first review the construction of GRFs, with an emphasis on binary classification.
We then describe the structure of XOR-like features and the reasons GDTs fail to capture those. 
We subsequently introduce an LRF implementation that recursively adds non-greedy layers as subtrees of depth $2$.
In the results section, we first train GRF and LRF binary classification models on synthetic data with XOR-like feature interdependencies. 
We find that the LRF outperformance is statistically significant as the signal-to-noise ratio becomes more opaque.  
Armed with these insights, we shift our focus to financial returns data, since financial markets are a hall-mark example of complex systems with low signal-to-noise ratio \cite{Patzelt2013,Anderson2018}.
LRF based predictions of daily signs of returns outperforms the ones of GRF across several asset classes. 
We then examine the copper futures price time series more closely and backtest a basic trading strategy in rolling windows from 2012 through 2020.
The LRF-based strategy's outperformance is supported by the observation of subtle, yet present, XOR-like feature interdependencies between short-term and long-term technical indicators.

\section*{Methods}
\subsection*{Random Forests with Greedy Decision Trees}
\label{sec:GreedySection}

We first provide an overview of how vanilla greedy random forests (GRFs) and their underlying greedy decision trees (GDTs) are built. 
For convenience, we limit our focus to binary classification.
Classification for a larger number of classes and regression variations is similar \cite{James2013}. 
For the remainder of this article, we label the two classes as ``$+$''  and ``$-$."
In the context of price-movement and the included examples, ``$+$''  represents a future upward movement in price and ``$-$" represents a future downward movement in price.
The reader familiar with GRFs can skip this part of the Methods and continue with the lookahead decision trees below. 

A RF is trained on a training dataset of $N$ samples, $\left\{ X_i \right\}_{i=1}^N$, where the datapoint $X_i$  is a vector of length $k$ , representing the $k$ different features that the algorithm may use for classification. 
We write $X_i = \left( F^{(0)}_i, F^{(1)}_i, \ldots, F^{(k-1)}_i \right)$ to describe the individual feature values $F^{(j)}_i$ of the vector $X_i$. 
To each datapoint in $X_i$ corresponds a (binary) target value $y_i$ (either $``+"$ or $``-"$) used to train the algorithm.
The RF is made up of a pre-specified number of $T$ decision trees that recursively split training samples with the objective of disentangling the two classes.
There are two important characteristics of a RF that ensure the creation of heterogeneous DTs. 
First, for each of the $T$ trees, we sample $N$ datapoints from $\left\{ X_i \right\}_{i=1}^N$ with replacement, such that each tree trains on a slightly different version of the data, still of size $N$. 
Second, at any split point, only a randomly selected subset of $\tilde{k} \leqslant k$ features is considered for the split. 
Common choices are $\tilde{k}=\sqrt{k}$ or $\tilde{k}=\log_2k$.

Conventionally, DTs are built through recursive binary splitting, often referred to as a greedy approach \cite{Breiman1984}. 
To start, the randomly selected training samples are initially contained in the ``root node" at the top of the DT, as can be seen in Figure \ref{fig:LRF_example}.
From the root node a split forms two tree branches and divides the samples into two child nodes.
This split is defined by one of the sample features and a specific value for that feature.  
The samples that were originally in the root node are separated based on whether their values for that feature are greater or less than the indicated split-value. 
The root node chooses this feature and value to minimize the classification impurity of the samples that will be divided into each of its two separate child nodes.
We use the Gini-score to measure classification impurity.
Using other impurity measurements like entropy is similar.  
For binary classification, with the classes ``$+$''  and ``$-$," the Gini-score, $G$, of a node is defined as:
\begin{equation}
	\label{eq:GiniScore}
	G = 2P_+(1-P_+) = 2P_-(1-P_-)    
\end{equation}
where $P_+$ and $P_-$ are the respective proportions of ``$+$" and ``$-$" observations present in the given node,
\begin{equation}
	P_+ =  1-P_- = \frac{\text{num.``$+$" samples in node}}{\text{total num. samples in node}}.
\end{equation}
The DT construction is described as greedy because each split is determined by optimizing the classification of the immediate split without taking into consideration the implications for future splits. 
That is to say, a greedy approach dictates that a parent split-node $S_0$ {is decided based purely on what is locally optimal.
In the context of the vanilla RF implementation considered here,} this means that one feature is selected from the available $\tilde{k}$ features, 
with a corresponding split-value $V_0$ so that the sample-weighted average Gini-score ${G_W}$ is minimized.
The Gini-score $G_W$ is defined as
\begin{equation} 
	\label{eq:GreedyWeightedGini}
	G_W = \frac {n_1G_1 +  n_2G_2} {n_1+n_2}  
\end{equation}
where the two child split-nodes $S_1$ and $S_2$ have respective Gini-scores $G_1$ and $G_2$, and $n_1$ and $n_2$ represent the number of samples contained in each. 
This process is then repeated for $S_1$ and $S_2$ and each of their subsequent child nodes until all observations contained in a node are of the same class or a stopping criteria, such as a tree depth threshold or a minimum number of required samples, is reached.
At this point, each of these terminal nodes, or ``leaves", represents a distinct, non-overlapping region of the predictor space.  

Once trained, a RF model can be used to predict class labels (``$+$''  or ``$-$") for new datapoints.
The RF's aggregate prediction is based upon the predictions of each of its DTs.  
To predict the classification of a datapoint with a single DT, one starts at the trained DT's root node.
One then navigates the DT by comparing the datapoint's feature values to those that define each split and then by following the DT's branches accordingly.  
Once a leaf is reached, the leaf's $P_+$ is interpreted as the signal strength, or ``probabilistic prediction'' that this datapoint belongs to the ``$+$'' class (and similarly for $P_-$). 
The entire RF's probabilistic prediction $P_+$ is the simply the average of the probabilistic predictions of all of its DTs. 
For binary classification tasks, the sample is  classified  as ``$+$'' if $P_+  > 0.5$ and as ``$-$'' otherwise. 
These probabilistic RF predictions are the basis for the trading strategies presented below. 
The idea to subsample the data-set, train on each subsample and then aggregate the individual predictions is known as {bagging} (short for bootstrap-aggregating). 
Bagging is one of the most effective computationally intensive procedures to improve on unstable estimates, especially for problems of high complexity \cite{Wager2014,Biau2016}. 
However, bagging cannot improve the accuracy of poor classifiers. 
If the individual learners are poor classifiers majority voting will still perform poorly (although with lower variance) \cite{dePrado2018}. 
It is thus particularly important to assert that the individual classifiers are individually capable.
This further motivates our introduction of step-wise lookahead DTs introduced below.

\subsection*{The XOR-Junction} 
\label{sec:XOR}

An ``exclusive disjunction" (XOR) is a logical operation that returns True when its two inputs differ (one True, one False) and False when they are the same (both True or both False).
A stylized XOR-like feature-interdependency between two features $F^{(0)}$ and $F^{(1)}$ is depicted in Figure \ref{fig:LRF_example} (b) for $100$ sample points.  
Both $F^{(0)}$ and $F^{(1)}$ have values that range from 0 to 1.  
All samples which have both $F^{(0)}, F^{(1)} \lessgtr  0.5$ belong to the ``$-$''class and otherwise belong to the ``$+$'' class. 

Ideally, a DT of depth 2 would recognize this XOR structure. 
It should determine that to best separate the data into two classes, $F^{(0)}$ should have a split value $V_0=0.5$ and $F^{(1)}$ should have a split value $V_1=0.5$.
However, if the DT attempts to determine the best split value for each feature in isolation of the other, it will likely not arrive at this result. 
For example, as shown in Figure \ref{fig:LRF_example}, the DT could split on $F^{(0)}$ at $V_0=0.25$, $V_0 = 0.5$, or $V_0= 0.8$.
In all these cases, the ratio of ``$+$" samples to ``$-$" samples, and hence the  Gini-score \eqref{eq:GreedyWeightedGini}, remains (on average) the same within both of the distinct regions.
This would likewise be true if the DT were to first split on $F^{(1)}$.
Myopically, the greedy DT discovers no significant classification advantage by splitting at any particular split value for either $F^{(0)}$ or $F^{(1)}$ and consequentially selects a globally suboptimal split. 

Therefore, a single GDT fails to capture such XOR-interdependencies. 
While the GRF may sometimes overcome this issue by averaging out the different, suboptimal, predictions, it will likely fail to do so in the presence of other features which have any explanatory power.
In such situations, it may be that hardly any of the trees will split on $F^{(0)}$ and $F^{(1)}$,  
as they will instead elect to split on these other less-explanatory features.
To address this shortcoming, we present an alternative -- stepwise lookahead, i.e. ``less myopic'' DTs which simultaneously optimize on three split-points, allowing one to better capture such subtle XOR-structures. 

XOR-type relationships are not just artificial constructions but do often play important roles in real problems. 
One example of this is the relationship between a country's possible fiscal policies and the resulting stability and resiliency of its economy.  
There are two possible sets of fiscal rules that can be imposed: (A) strict debt and budget balance rules, and (B) strict expenditure and revenue rules.
To achieve a stable economic regime, a country can implement either (A) \textit{or} (B), but not both.
If both (A) \textit{and} (B) are implemented, the excessive fiscal regulation can impede a country's ability to react to economic turbulence and will result in an unstable economic regime \cite{Hackett2015}.
Biology gives us another example.
The survival of the Drosophila (fruit fly) depends on its gender and the \textit{Sxl} gene activity \cite{Page2003}. 
Females only survive if the gene is active, whereas males only survive if it is not. 
We find yet another example below in our examination of copper futures price data which shows that short-term and long-term technical indicators also form such XOR-junctions.
The existence of these patterns makes the ability to infer such interdependencies directly from data vital for modern data science. 
 
Binary XOR-junctions are of course not the only type of relationships for which the GDT is suboptimal by construction. 
It is straight forward to think of a the generalization to the so-called $k$-XOR problem, which consists sequential application of $k-1$ XOR operations. 
Any DT that looks ahead less than $k$ depths will fail to adequately resolve such interdependencies \cite{Esmeir2004}. 
While we focus on binary XOR-junctions in the remainder of this article, our results are expected to naturally generalize to other such patterns, at the expense of higher computational cost.

\subsection*{Random Forest with Stepwise Lookahead Decision Trees}
\label{sec:LookaheadDT}

The near-sighted, greedy approach to DT construction only focuses on the two immediate child nodes to determine a split.
This may lead to suboptimal decision trees, most notably for XOR-like pair-wise feature interdependencies shown in Figure \ref{fig:LRF_example} (b). 
Many different approaches have been undertaken to implement more optimal DTs \cite{Bennett1994,Uney2006,Norouzi2015,Barros2015,Bertsimas2017,Firat2018,Verwer2019,Blanquero2020}. 
The computational complexity of training non-greedy DTs however grows exponentially with the number of nodes, as opposed to linearly for greedy ones.  
Other proposed implementations have attempted to deal with this NP-hard problem through approximations and simplifications that range from 
iterative linear programming \cite{Bennett1994} to stochastic gradient descent \cite{Norouzi2015} and mixed integer programming \cite{Uney2006}. 
See Verwer and Zhang \cite{Verwer2019} for a detailed technical account of these approaches. 

Albeit better in theory, lookahead DTs (LDTs) possess a higher risk of overfitting, which is of particular concern when dealing with low signal-to-noise ratio problems like those discussed below. 
This issue might also be in part responsible for the lacking empirical support of these previously proposed methods \cite{Murthy1995,Esmeir2004,Rokach2016}.
Here, we instead seek to strike a balance between computational complexity, risk of overfitting, as well as practicability. 
We achieve this by constructing DTs with a lookahead of only depth $2$. 
Since this means we only optimize across three split nodes at a time, a brute force optimization remains feasible, while also keeping the risk of overfitting lower than with globally-optimized LDTs. 
From a practical perspective, the two-step lookahead DT allows us to capture binary feature interactions better than GDTs. 
However, it is worth noting that this method will likely fail to capture specific higher order $k$-nary interaction patterns. 
If such higher-order feature interdependencies are suspected to be presented in a given dataset, one should resort to approximate, higher order techniques \cite{Bennett1994,Uney2006,Norouzi2015,Barros2015,Bertsimas2017,Firat2018,Verwer2019,Blanquero2020}.
The LDTs that we propose are not limited to depth $2$, but can be further expanded to greater (even-sized) depths by recursively appending non-greedy subtrees of depth $2$, 
similar to how a GDT appends individual nodes. 
Nevertheless, we find that for practical applications to problems with low signal-to-noise ratio, LRFs with trees of depth $2$ outperform those with deeper trees. 

Each non-greedy DT (or subtree) of depth 2 has the structure shown in Figure \ref{fig:LRF_example} (a).
It consists of 3 split-nodes that are jointly determined. 
As in greedy construction, each split node only sees a random subset of $\tilde{k} \leqslant k$ features, giving rise to at most $\tilde{k}^3$ different feature combinations (or less due to symmetries). 
To determine the split-value of any given feature, we discretize the feature's value range into $B \leqslant N$ equal-sized buckets based on the sample quantiles to avoid susceptibility to outliers. 
Out of these $B~\tilde{k}^3$ possible DT structures, we select the one that minimizes the cumulative sample-weighted Gini-score
 \begin{equation}  	
	G_C =  \sum_{i=0}^{i=3} n_i~G_i
	\label{eq:NGTotalGini}
\end{equation}
where $n_i$ is the number of samples and $G_i$ is the Gini-score \eqref{eq:GiniScore} of leaf $L_i$. 
If the maximum depth of the DT is greater than $2$, the above process is repeated recursively to grow subtrees in steps of depth 2 until a stopping criteria is met (not enough datapoints in a leaf, all datapoints in a leaf belong to the same class, etc.).
Apart from the construction of its decision trees, a LRF functions the same way a GRF does.  
It likewise makes classification predictions by averaging the probabilistic class predictions across its LDTs.

Finally, one more note on the aspects of computational complexity - the larger $B$ is, the more granular the resolution of potential splits. 
The upper bound is at $B = N$, in which each feature value is unique and considered as a potential split point. 
The presence of categorial features immediately reduce that number. 
In the results section, we have varied $B$ from $30$ to $N$ without a qualitative change in our results. 
It can also be handled as a {hyperparameter} and optimized via $k$-fold cross-validation. 
Even for applications where $k$ reaches the hundreds, and $N$ is in the thousands, the evaluation of all $B~\tilde{k}^3$ decision trees remains computationally feasible for standard modern computers.

\subsection*{Benchmark Methods and Selection of Hyperparameters}
\label{sec:meta}

In this article, we assess the effectiveness of LDTs versus GDTs.
Rather than focus on the performance of the DTs in isolation, we train multiple DTs as an ensemble in random forests (LRF vs. GRF). 
We compare the performance of our LRF against a standard implementation of the GRF, 
namely Sci-Kit Learn's RandomForestClassifier \cite{Pedregosa2011} which is built on an optimized version of the CART algorithm \cite{Breiman1984}. 

Both LRF and GRF train their DTs on bootstrapped data and then average the individual predictions, so-called bagging. 
This way, the mistakes of a single DT are compensated for by other decision trees in the forest \cite{Rokach2016}.
As such, random forests are known to be particularly robust with respect to noise \cite{Breiman2001}, which makes them ideal candidates for our application to financial data. 
However, bagging cannot improve the accuracy of poor classifiers. 
In contrast, individual classifiers can also be trained in sequence on residuals, an approach known as boosting \cite{Schapire1990}. 
Compared to bagging, boosting methods may actually improve on the prediction accuracy of individually weak classifiers \cite{Freund1996}.
As such, it is thus possible that less optimal GDTs, when boosted, achieve greater prediction than the individual classifiers could. 
Therefore, we will also compare the performance of our LRF against boosted GDTs. 
{Specifically, we use XGBoost \cite{Chen2016}, from hereon abbreviated as XGB.}
That said, it is of course also possible to use the LDT presented in this article and train them sequentially for boosting. 
This is, however, beyond the scope of this article. 

In summary, we will compare the performance of our stepwise LRF 
against a vanilla GRF, 
against an individual, pruned GDT, 
and against boosted DTs. 

Each of these methods requires the specification of a set of {hyperparameters}.  
The most important parameters are common to all these methods. 
Namely, 
the maximum depth of each tree, 
the minimum number of samples in each final leaf, 
and the fraction of randomly selected features each split node gets to see. 
We determine the optimal {hyperparameters} by means of  $k$-fold cross-validation. 
Detailed information on the {hyperparameters} for each algorithm are found in the SI Appendix. 
\footnote{The SI Appendix is available from the authors upon request.}

\subsection*{Tuning the Signal-to-Noise Ratio with Synthetic XOR Data}
\label{sec:XOR}

We next systematically examine the relationship between signal-to-noise ratios, XOR-patterns, and the associated performance of different classifiers. 
It is difficult to evaluate the relationship between data complexity and classifier performance, since the concept of data complexity is illusive itself \cite{Ho2002}. 
But to do so, we create $N=2,000$ synthetic data points  \`a la Figure \ref{fig:LRF_example} (b), 
allowing us to control the experimental setup and examine the classifiers' relative performance across varying conditions.  

We generate the values for the first two features, $F^{(0)}$ and $F^{(1)}$, by sampling for each $N$ times uniformly from the interval $[0,1]$. 
We denote by $F^{(0)}_i$ the $i$-th such data-point for feature $0$ $(i=1, \ldots, N)$, and similar for feature $1$. 
By definition, $F^{(0)}$ and $F^{(1)}$ are uncorrelated. 
Denote by $B = B(\rho)$ the realization of a Bernoulli-distributed random variable with $\mathbb{P} \left[  B = 1 \right] = \rho$. 
The binary target values $\{ y_i \}_{i=1}^N$ are encoded by setting ``$+$'' to $1$ and ``$-$'' to $0$. 
In accordance with Figure \ref{fig:LRF_example} (b), the $i$-th target variable is then generated as 
\begin{equation}
	y_i (\rho)= \left\{ 
			\begin{array}{cc} 
				B(\rho)     &\text{ if } \Theta \left( F^{(0)}_i \right)  \neq \Theta \left( F^{(1)}_i \right)  \\
	 			1-B(\rho)  &\text{ if } \Theta \left( F^{(0)}_i \right)  =      \Theta \left( F^{(1)}_i \right) 
			\end{array} 
			\right. 
		\label{eq:y(rho)}
\end{equation}
where we have defined the (shifted) Heaviside function $\Theta$ as $\Theta(x) = 0$ if $x < 1/2$ and $\Theta(x) = 1$ if $x \geqslant 1/2$. 
The parameter $\rho$ allows us to tune the signal-to-noise ratio for the XOR-pattern from pure signal and zero noise $(\rho=1)$ down to zero signal and pure noise $(\rho=1/2)$.
In the results section, we test the prediction accuracy of LRF and GRF as a function of $\rho$, and find that the outperformance of the LRF is particularly pronounced in the area where $\rho$ approaches $1/2$.

\section*{Results}

\begin{center}
\begin{figure}[h!]
   \includegraphics[width=\linewidth ]{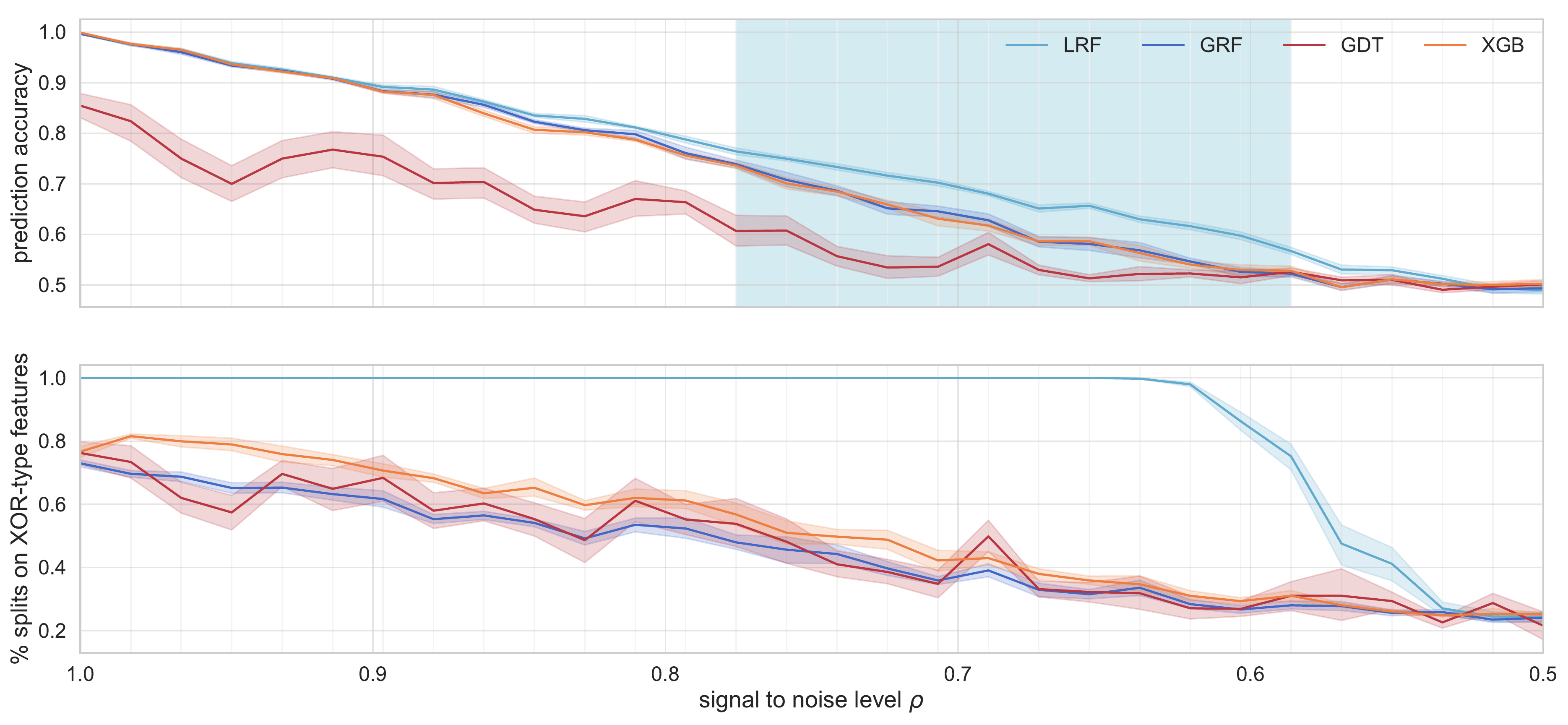}
   \caption{ 
The top plot shows the prediction accuracy for different classifiers as a function of the signal-to-noise level $\rho$ defined in \eqref{eq:y(rho)}. 
Towards the left side (signal dominates) and the right side (pure noise), the lookahead random forest is not distinguishably better than greedy variants (GDT or XGB). 
However, there is a transition regime (blue shaded background) of lower signal to noise ratio where the LRF outperforms since the GDTs that underly the other algorithms fail to regularly identify the relevant features. 
This is observed in the bottom plot, where, on the left y-axis. the relative importance of the individual features is shown.
The LRF correctly identifies the XOR-like features much more consistently than the other methods, thereby accounting for its outperformance. 
The individual GDT underperforms all of the ensemble methods. 
}
\label{fig:synthetic_data}
\end{figure}
\end{center}

\subsection*{Performance as a Function of the Signal-to-Noise Ratio}
\label{sec:XOR}

At least in theory, the (two-step) lookahead DTs ought to better capture (binary) XOR-like feature interdependencies compared to their greedy counterparts. 
But why are empirical results \cite{Murthy1995,Esmeir2004,Rokach2016} mixed? 
One reason may be that such XOR-like feature interdependencies are rarely observed in practical applications. 
However, we will encounter such interdependencies below when we analyze the interaction between short-term and long-term technical indicators for the prediction of copper futures returns. 
Let us thus assume we are dealing with a dataset in which such XOR-like feature interdependencies are present. 
Is it then really the case that the LDTs outperform GDTs? 

We investigate this question by training and predicting on synthetic datapoints for which we can set the value of $\rho$ which parameterizes the signal to noise level. 
For a fixed $\rho$, we sample $N=2,000$ datapoints as follows: 
We work with $k=8$ features of which the first two features, $F^{(0)}$ and $F^{(1)}$, are related to the target variables as described in equation \eqref{eq:y(rho)}. 
The remaining six features $F^{(2)}, \ldots, F^{(7)}$ are uncorrelated with $y$ and randomly distributed in the interval $[0,1]$.  
They serve as obfuscating random noise. 
Given this synthetic data we train on 75\% of the data and use the remaining 25\% for out of sample classification. 
The following four binary classifiers are trained: LRF, GRF, GDT and {XGB}.
{Hyperparameters} are selected by means of $5$-fold cross-validation. 
See Methods for details. 

We measure performance on the test data by the accuracy score, i.e. the fraction of correct predictions out of all predictions. 
For fixed $\rho$, this results in one accuracy score per binary classifier. 
As a robustness check, we repeat the above experiment $M=20$ times, to get $M$ different prediction accuracies each. 
We then use the average across those values as representative prediction accuracy and the standard deviation as error bars. 
The outcome of this experiment is depicted in Figure \ref{fig:synthetic_data}. 
As anticipated, on the far right, where the features are dominated by pure noise $(\rho \gtrsim 0.5)$, all prediction accuracies hover around the level of random guessing ($50\%$ accuracy). 
Perhaps more surprisingly is the left side of the plot where the signal dominates $(\rho \lesssim 1)$. 
With the exception of the single decision tree, here, the outperformance of the LRF over greedy methods is marginal, if at all existent. 
By contrast, for intermediate values, and especially those closer to the pure noise separation line, the LRF significantly outperforms all other methods. 

To better understand these observations { and the respective contributions of the different features, we conduct a relative feature count across the entire forest.
The relative importance of feature $F^{(i)}$ is then calculated as the number of times the forest splits on feature $F^{(i)}$ divided by the total number of split points.
} 
Although more statistically sophisticated methods exist \cite{Kursa2010,Lundberg2020}, here we rely on a basic feature count across all split-nodes, 
since it allows for a straight-forward interpretation and comparison across different tree-based methods. 
The bottom plot in Figure \ref{fig:synthetic_data} shows that the LRF picks up on the XOR-features significantly more regularly than all other methods. 
Indeed, the LRF cross-validation consistently selects $\tilde{k} = k$, i.e. that all features are visible to any node, in order to correctly split on the XOR-feature. 
By contrast, the greedy methods are frequently mislead by the noise, such that the XOR features are not always selected to split on. 

In practical applications, it is unrealistic that we will work with such a dataset where many of the considered features are purely noisy, especially if thorough feature selection \cite{Kursa2010} has been performed prior.
Therefore, we have extended the above tests whereby we have replaced the $6$ random noise features with features that are weakly linearly related to the target variable. 
{
By construction, these linear features are less explanatory than the XOR-type features and therefore should ideally not be selected over the combination of the XOR-type features.
Hence, we observe that in this case the results are similar, but that the replacement of the noise with the linear features does slightly improve the prediction accuracy of the GRF, GDT, and XGB models.  
While the LRF's outperformance is reduced, it is still pronounced and supports the choice of the LRF model over the others for such a dataset. 
}
When XOR-dependencies are removed, the performance among the LRF, GRF and {XGB} are similar. 
The outcomes of these results are detailed in the SI Appendix. 

\subsection*{Application to the Prediction of Financial Returns}
\label{sec:application} 

Above, we have shown that the LRF outperforms more greedy variants in the presence of XOR-like feature interdependencies as well as unfavorable signal-to-noise conditions. 
Here, we turn our attention to real datasets to test if outperformance of our LRF variant is observed in practical applications. 
{The UCI data sets are often considered for such use cases.
However, Esmeir and Markovitch \cite{Esmeir2004} have already examined 12 UCI datasets based on a two-step lookahead decision tree. 
They found that, conditional on complex feature interdependencies being present, the lookahead tree outperforms pruned, greedy trees. 
This previous work, alongside our insights from the synthetic dataset, indicate that UCI data is not a suitable testing ground for our consideration.
Moreover, across UCI data sets, with the exception of the very few that are unbalanced} (with {heavily} populated majority classes), the best {achieved} accuracy scores are always much higher than the fraction of the majority class \cite{Fernandez2014}. 
{In contrast, we expect noisy environments, with relatively low expected prediction accuracies, to be where the LRF primarily outperforms.}
We thus turn our attention to financial markets. 
Financial markets are known to exhibit a particularly unfavorable signal-to-noise ratio and future price returns are thus notoriously hard to predict \cite{dePrado2018}. 
If easily recognizable signals do arise they are fleeting and quickly exploited.
Active traders rapidly spot and act on such opportunities, pushing prices in their predicted directions. 
In other words, often times the very existence of a signal gives rise to its own decay \cite{Mclean2016}. 

It is amusing to think of an analogy from image recognition. 
Assume one is training a classifier to recognize dogs.  
Imagine that these dogs notice they are being classified as dogs, and subsequently alter their appearances.
Clearly an absurd scenario, but this is very much the reality in finance. 
This constant race to the bottom has recently given rise to the application of generative adversarial networks which identify the states and portfolios with the most unexplained pricing information \cite{Chen2019}. 

Accounts of machine learning applications in finance are manifold, see for instance ref. \cite{dePrado2018,Dixon2020} for a recent overview. 
In many domains of machine-learning, for instance image recognition, research is mostly concerned with improving on already impressive prediction accuracies. 
By contrast, in finance, the question is often whether machine learning methods are able to predict financial returns at all \cite{Mclean2016,Bailey2016,Chen2019}.
Such low signal environments make for the ideal testing ground for our LRF. 

With the same methodology as for the synthetic data above, we have trained LRF, GRF, GDT and {XGB} to predict the signs of daily financial returns. 
We have tested the prediction across 16 different assets. 
The challenge is that there is a myriad of assets and asset classes that can be tested.
We have therefore narrowed down the eligible universe by picking assets that are at the sweet spot between being liquid enough to trade without impact, 
but not so liquid that return prediction based on simple technical indicators seems out of reach due to market efficiency.
The approach towards asset selection is detailed in the SI Appendix.
The selected 16 assets are all commodity rolled front month future contracts, shown in Table I. 

As features, we have used eight basic technical indicators derived from the asset's open, high, low, close and volume price data: 5-day and 20-day RSI, 5-day and 20-day Z-scored volume, 5-day and 20-day return sign correlation, overnight-gap and close location value. 
A detailed description of these indicators is found in the SI Appendix. 
Granted, these indicators are very crude and the industry professional is advised to rely on a more sophisticated selection of features. 
But here, our main concern is to test the relative, rather than absolute, performance of the LRF compared to other methods.

Following the same methodology applied to the synthetic data above, we have used $k$-fold cross-validation to determine the parameters for LRF, GRF, GDT and {XGB}.
We have trained all models on prices from 2012 to 2017 and predicted out of sample prices from 2018 to 2020. 
The prediction target is the sign of the next day's return (positive or negative). 
Table I shows the classification accuracies across 16 commodities for the four classifiers. 
For the reasons elaborated above, most of these predictions don't deviate much from the baseline accuracy of always predicting the majority class (\% majority in Table I). 
A binomial test with a $p$-value at the $5\%$ confidence level (denoted by an asterisk) indicates that the outperformance was statistically significant. 
The actual $p$-values are also reported in Table I. 
Not surprisingly, neither of the methods outperforms consistently across asset classes.
At most, this is only partially to be blamed on the methods, but more so on the crude indicators that have been used. 
However, we notice that, at the $5\%$ confidence level, the LRF outperforms on 5 out of 16 assets, followed by the GRF which outperform on only 3 out of the same 16 assets. 
Single GDT and {XGB} only outperform once, each. 
We will now investigate more whether such outperformance is actually to be explained by the presence of XOR features.

\begin{figure}[!htb]
	\centering
	\includegraphics[width=\textwidth]{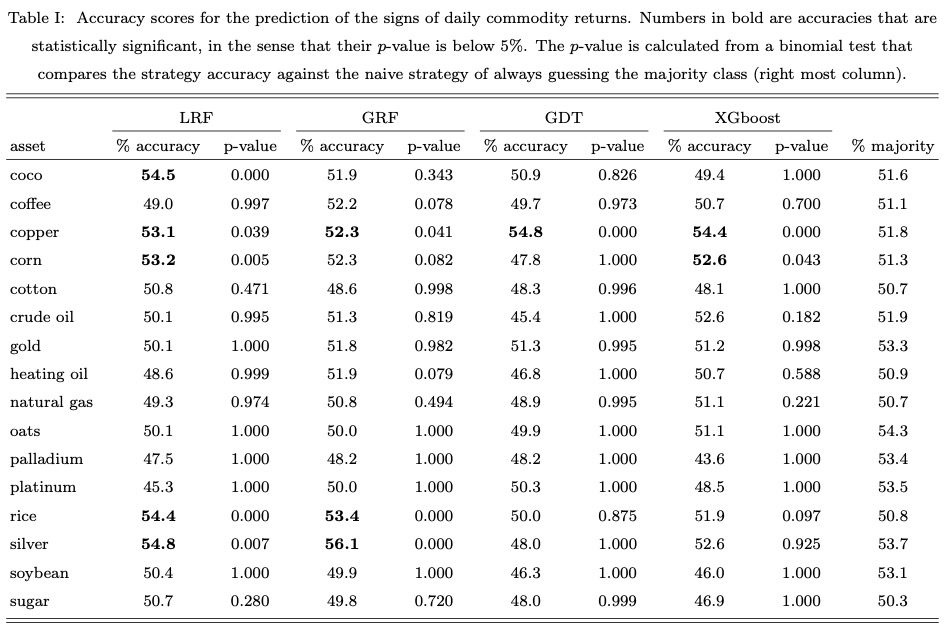}
\end{figure}

\subsection*{XOR-Relationships Amongst Technical Financial Indicators - The Example of Copper Futures}
\label{BTCSection} 

The above analysis across different financial assets has shown that the LRF outperforms more greedy measures on some, but not all assets. 
As our tests on synthetic data have highlighted, outperformance can be expected when XOR-like feature interdependencies are present. 
In this section, we single out a single asset, copper futures.
We have selected copper because it is the only asset for which a statistically significant outperformance of the benchmark - prediction accuracy for always predicting the majority class- is observed across the classifiers (see Table I). 
We examine in greater depth the XOR-like feature interactions present within the copper futures dataset and how these translate to higher prediction accuracies with LRFs.
Here, we only focus on LRFs compared to GRFs because the individual GDT and {XGB} models consistently underperform across the 16 tested assets. 

To make explicit the real consequences of such outperformance, we extract a concrete trading strategy from the trained binary RF classifier that predicts the sign of daily copper returns. 
Prior to each (out of sample) trading day, the RF's $2,000$ trained DT's  return a binary prediction - whether the following day's price return will be positive or negative. 
If, for instance, $1,400$ of these DTs predict that tomorrow's return will be positive, this translates to a signal strength of $1,400/2,000 = 70\%$, and so forth. 
For the following trading day, we are long/short the asset if the signal is $\theta \%$ above/below the neutral $50\%$.
If it is not, we take no position for that trading day.
The threshold $\theta$ is a {hyperparameter} to be selected along with the classifier {hyperparameters}. 
Performance is evaluate by means of the Sharpe ratio \cite{Lo2002}.
We repeat the above process in rolling windows in steps of 75 trading days so that we train on the most recent data. 
The same eight basic technical indicators previously introduced are used. 
We further assume instantaneous trade execution and do not account for slippage or trading costs.
These simplifications do not affect the analysis presented, 
since our objective is to evaluate the relative effectiveness of different classifiers, not to build out viable trading strategies. 
Details are found in the SI Appendix.

\begin{center}
\begin{figure*}[!htb]
   \includegraphics[width=\linewidth ]{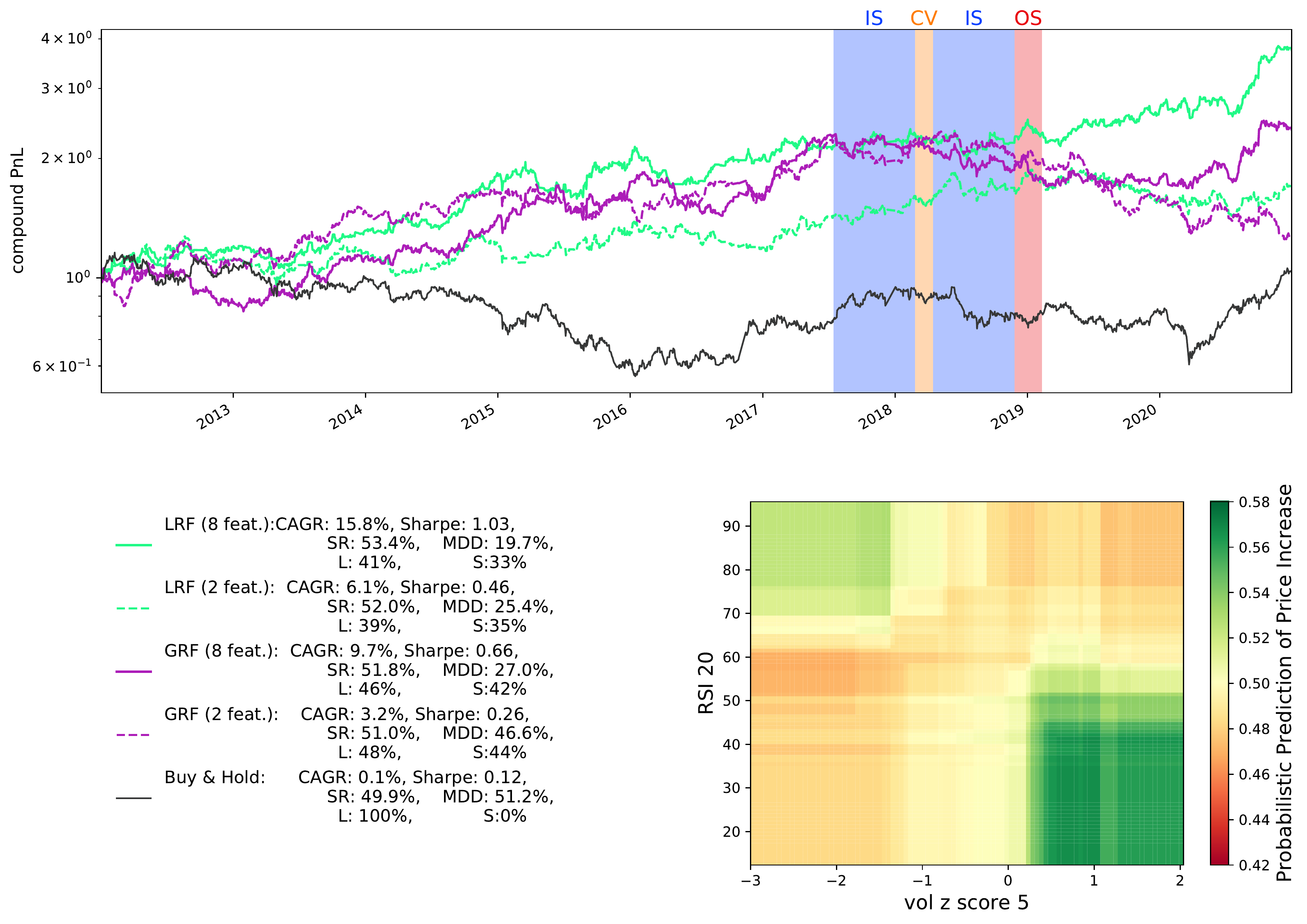}
   \caption{ 
   The top plot includes the cumulative returns from 2012 through 2020 for the different trading strategies. 
   The black line represents the benchmark ``buy and hold'' strategy for copper futures.
   The solid green (purple) line represents the returns of the {optimized} LRF (GRF) trading strategy detailed in SI Appendix, based on the 8 technical indicators detailed in SI Appendix. 
   The legend shows the average cumulative annual growth rate (CAGR), annualized Sharpe ratio, success rate (SR), maximum drawdown (MDD), and the fraction of time that a long (L) or short (S) positions is held. 
   The LRF clearly outperforms the GRF across various metrics. 
   The shaded background colors depict one of the rolling windows for in-sample (IS) training, cross-validation (CV) and out-of-sample (OS) prediction. 
   These windows are rolled in steps of 75 (trading) days. 
   The bottom right heatmap visualizes the ``probabilistic'' classification (fraction of DTs that predict at positive return) of a LRF trained on just two features (cf. axis labels). 
   This structure is akin to the XOR-pattern of Figure \ref{fig:LRF_example}. 
   The dashed green (purple) line represents the returns of the optimizes LRF (GRF) trading strategy detailed in SI Appendix, based on only the two XOR-like technical indicators. 
   The outperformance of the LRF is yet more pronounced in this case.  
    }
   \label{fig:Cu_strategy}
\end{figure*}
\end{center}

The outcome of this strategy is depicted in Figure \ref{fig:Cu_strategy}. 
From 2012 through 2020, the benchmark 100\% long "buy \& hold" strategy has an annualized Sharpe ratio of 0.12 and an average annualized return of 0.1\%.  
Over that same period, the GRF-based strategy achieves an annualized Sharpe ratio of 0.66 and an average annualized return of 9.7\%.
The LRF outperforms both of these benchmarks with an annualized Sharpe ratio of 1.03 and an average annualized return of 15.8\% while staying out of the market more frequently. 
More performance measures are summarized in Figure \ref{fig:Cu_strategy}.

To better understand the LRF outperformance, we investigate the feature combinations that best explain the classification of the dataset.
Upon examination of the DTs used in the LRF-based strategy, we identify the five most frequently selected feature combinations that are used by the nodes to split the data.
For each of these five feature combinations, we train an LRF on all of the available data and visualize its probabilistic prediction using heat-maps.
We only consider the first batch of in-sample data to avoid a subsequent lookahead bias.
Amongst these, we find that the 5-day volume Z score and the 20-day RSI most demonstrate an XOR-like relationship, as is shown in the heatmap included in Figure \ref{fig:Cu_strategy}.
We can use this heatmap to attempt to understand how these features interact in a manner predictive of price movement.  
The 5-day volume-based Z-score indicates whether present trading volume is low or high relative to its short-term levels.  
The 20-day RSI indicates whether or not an asset is overbought (large RSI) or oversold (low RSI). 
When the 5-day volume Z score is large and the 20-day RSI is large, the probabilistic predictions are generally indicative of a price decrease (the orange top-right 'quadrant' of the heatmap).  
There exist multiple possible interpretations of this relationship, such as the following.
{When} the RSI is large, the asset is most likely overbought. 
At the same time, if trading volume tends to mean-revert, and is currently relatively high, it is expected to decrease in the near future. 
Together, this decrease in trading volume, coupled with the asset being overbought, suggests a price depreciation. 
On the other hand, if the RSI is large but volume is low, a price increase is expected. 
Deeper examination of this feature relationship as well as the other two `quadrants' is beyond the scope of this paper.
The key takeaway is that such XOR-like patterns are present in actual datasets.

The 5-day volume Z score and the 20-day RSI are now isolated as the lone features for GRF and LRF-based trading strategies. 
We only consider these $k=2$ features and set  $\tilde{k}=k=2$ so that both features can be considered at each split point.  
The strategies are also constructed according to the previously-mentioned rolling windows process and their returns are also included in Figure \ref{fig:Cu_strategy}.
While, unsurprisingly, the performances of both these strategies suffer as compared to those of the 8-feature versions, the disparity between the LRF and GRF models' performances is dramatically accentuated.  
{
While the total compound returns of the two strategies are similar (observed by the fact that both dashed lines finish at similar levels), 
the annualized, and risk-adjusted performance measures of the GRF strategy are much better. 
The GRF-based strategy Sharpe ratio shrinks to 0.26, while the LRF based strategy is still able to achieve a Sharpe ratio of 0.46, which is almost double. 
Similarly, the draw-down of the LRF strategy is half of the GRF one. 
}
This suggests that LDTs effectively account for the XOR-like relationship while the GDTs do not.

{
As we show in the SI Appendix, 
the above conclusions remain unchanged when also considering XGB and GDT models. 
Both methods underperform the LRF on both the eight and two features versions. 
The performance of the LDT on eight features is, not surprisingly, underwhelming, considering that the tree depth is limited to at most $4$. 
Once restricted to only two features, the LDT outperforms all other methods except the LRF, which it still underperforms. 
We interpret this as a result of the bootstrapping, which is a useful property of decision tree forests that decreases variance and accordingly smoothens the trading signal. 
In other words, decision forests are to be preferred over decision trees in high-noise environments. 
} 

\section*{Discussion}

Tree based methods are popular in data science due to their many advantages such as interpretability and robustness to noise. 
Most applications rely on suboptimal decision trees (DTs) with greedy top-down induction, mainly for computational convenience but also because globally optimized DTs failed to convince in general \cite{Rokach2016}.
In this article, we have shone light on this issue by checking the conditions under which ensembles of lookahead DTs (LDTs) actually outperform their greedy counterparts. 
To this end, the canonical example of the XOR-junction,  shown in Figure \ref{fig:LRF_example} (b), was considered. 
In general, a greedy DT (GDT) is expected to fail to find the optimal split-points that accurately capture XOR-like interdependencies, whereas an LDT that optimizes across multiple split-points can. 
We have tested this hypothesis on synthetic data for which we can systematically tune the signal-to-noise ratio (SNR). 
Perhaps not immediately intuitively, we found that the outperformance is significant only in a regime where the the SNR is low (top plot in Figure \ref{fig:synthetic_data}). 
By its construction, a greedy DT structure is more easily misled by noise. 
But when considered as an ensemble, the individual errors of the GDTs average out and the GRF's predictiveness is retained at high signal levels.
However, as the relative noise grows, this no longer remains the case and the GRF's predictiveness declines.
We have shown that the LDT is indeed better at identifying the relevant XOR-like features amidst noise and accordingly, the LRF does not experience as rapid of a performance deterioration as the GRF (bottom plot in Figure \ref{fig:synthetic_data}).

Given this insight that outperformance is especially prevalent in high-noise environments, where the danger of overfitting is large, we have then focused on stepwise lookahead DTs in tiers of depth $2$.  
This was further motivated by the fact that $k$-nary XOR-like feature interactions with $k > 2$ are theoretically possible, but surmized less relevant for practical applications. 
As such, the two-step lookahead DT that we present strike a good balance between capturing binary feature interdependencies and affordable additional computational cost. 
Our implementation allows for trees with greater (even sized) depth by recursively appending additional non-greedy layers as subtrees of depth 2.    
However, our {hyperparameter} selection consistently selected LDTs of depth 2 as optimal. 
Indeed, greater depth quickly results in overfitting, especially in areas with low SNR. 

{
On a computational note, we have relied on a brute-force implementation of depth two trees, optimized across three split points. 
Since parallelization across trees is straight forward, this did not cause any computational problems. 
However, approximative optimization approaches such as gradient-decent \cite{Norouzi2015}, mixed-integer formulations \cite{Uney2006} or anytime inductions \cite{Esmeir2004} 
are expected to leave our conclusions qualitatively unchanged, 
but provide the additional benefit of faster execution. 
}

We have argued that financial returns are the ideal testing ground to test our method in practice, since finance has arguably one of the lowest SNR by construction. 
Any evident signal is immediately arbitraged. 
Looking for less common feature interdependencies that may be uncovered only by LDT, rather than GDT, is thus a promising new avenue. 
Accordingly, we have compared the prediction accuracy of binary returns for both greedy and lookahead methods across 16 assets. 
We found mixed evidence, with a slight tendency for lookahead methods to outperform. 
This result is not surprising, as the prediction of returns, or even just their signs, is known to be a particularly hard and irregular problem. 
In particular, better results could be expected by implementing more sophisticated and less scrutinized indictors, but this is not the purpose of our paper. 

To gain more insights into the basis for potential outperformance of lookahead methods, we have singled out a single asset, copper futures, and examined the classifiers' performance on it in greater depth. 
We have trained both lookahead random forests (LRFs) and vanilla greedy random forests (GRFs) and generated respective long-short trading strategies. 
The LRFs-based strategy was observed to outperform the GRFs-based strategy (top plot in Figure \ref{fig:Cu_strategy}), which suggests the presence of XOR-like feature interdependencies. 
By scanning the feature pairs that were frequently selected by the LRF, we have indeed found evidence for such XOR-like patterns. 
Specifically, we found that there is an interplay between short-term and long-term technical indicators (bottom right plot in Figure \ref{fig:Cu_strategy}). 
When isolating only these two features, the GRF drastically underperforms the LRF, which further highlights our main message that the detection of XOR-like features is sometimes crucial for practical applications. 

All our results have been benchmarked against some of the most common decision tree-based algorithms in use today, most notably the CART based greedy random forest and {XGB}.
We have refrained from testing a boosted version of stepwise LDTs as we have focused on particularly noisy datasets, where robustness against noise by means of bagging is more important. 
For practical applications, we highlight the largely neglected option of taking into consideration three split nodes at a time, rather than just a single one. 
When feature interdependences might exist within a given dataset, the two-step lookahead DT can be viewed as a valuable replacement for the GDT within ensemble methods, as it offers potential performance enhancement with little downside.

\section*{{Conclusions}}

{
Along with the increasing popularity of machine learning, decision tree based methods have been proven useful across a large range of applications. 
The most common implementation of these methods rely on sub-optimal, greedy constructions of these trees. 
Globally optimal, lookahead trees were proposed more than two decades ago, however, their mainstream adaptation has been lacking. 
In this article, we have questioned why this is the case and examined the applicability of such lookahead trees.
We have shown that non-linear feature interdependencies alone do not necessarily call for an application of lookahead trees. 
Instead, we have shown that such methods are particularly useful when both feature interdependencies and a low signal to noise ratio exist. 
Financial timeseries are typical examples of such datasets, which we have subsequently examined in more depth. 
We have shown that technical indicators can display XOR-like interdependencies, and that lookahead trees better unearth them. 
In conclusion, we now better understand why wide adoption of more optimal tree construction has been lacking -
it is not needed to achieve significant prediction accuracies for datasets with relatively favorable signal to noise levels.
However, when working with datasets that do not readily fall into this category, lookahead trees can provide important improvements. 
We hope that our results will spur further academic research and industry applications. 
In particular, we appeal to practitioners in the financial industry to consider lookahead decision trees when predicting patterns within financial timeseries. 
}

\bibliography{bibliography.bib}

\balance


\end{document}